\documentclass[11pt,a4paper]{article}
\usepackage{times,latexsym}
\usepackage{url}
\usepackage[T1]{fontenc}

\usepackage{enumitem}
\usepackage{multirow}
\usepackage{booktabs}
\usepackage{graphicx}
\usepackage[breakable]{tcolorbox}
\usepackage{wrapfig}
\usepackage{amsmath}
\usepackage{amsfonts}
\usepackage{subcaption}

\usepackage[acceptedWithA]{tacl2021v1}

\usepackage{xspace,mfirstuc,tabulary}

\newif\iftaclinstructions
\taclinstructionsfalse 
\iftaclinstructions
\renewcommand{\confidential}{}
\renewcommand{\anonsubtext}{(No author info supplied here, for consistency with
TACL-submission anonymization requirements)}
\newcommand{\instr}
\fi

\iftaclpubformat 

\else

\fi

\title{Goal Alignment in LLM-Based User Simulators for Conversational AI}

\author{
  \textbf{Shuhaib Mehri}$^1$ \quad
  \textbf{Xiaocheng Yang}$^1$ \quad
  \textbf{Takyoung Kim}$^1$ \quad
  \textbf{Gokhan Tur}$^1$ \quad
  \textbf{Shikib Mehri}$^2$ \quad \\
  \textbf{Dilek Hakkani-T\"ur}$^1$ \\[10pt]
  $^1$University of Illinois Urbana-Champaign, $^2$Contextual AI \\[5pt]
  \texttt{\{mehri2, dilek\}@illinois.edu}
}

\date{}

\begin{document}
\maketitle

\begin{abstract}
User simulators are essential to conversational AI, enabling scalable agent development and evaluation through simulated interactions. While current Large Language Models (LLMs) have advanced user simulation capabilities, we reveal that they struggle to consistently demonstrate goal-oriented behavior across multi-turn conversations, which is a critical limitation that compromises their reliability in downstream applications. We introduce User Goal State Tracking (UGST), a novel framework that tracks user goal progression throughout conversations. Leveraging UGST, we present a three-stage methodology for developing user simulators that can autonomously track goal progression and reason to generate goal-aligned responses. Moreover, we establish comprehensive evaluation metrics for measuring goal alignment in user simulators, and demonstrate that our approach yields substantial improvements across two benchmarks (MultiWOZ 2.4 and $\tau$-Bench). Our contributions address a critical gap in conversational AI and establish UGST as an essential framework for developing goal-aligned user simulators. All code and data is released to facilitate future research \footnotemark[1].
\end{abstract}

\footnotetext[1]{\url{https://github.com/Shuhaibm/user_simulator_goal_alignment}}
\section{Introduction}

\begin{figure}[htbp]
    \centering
    \includegraphics[width=0.9\linewidth]{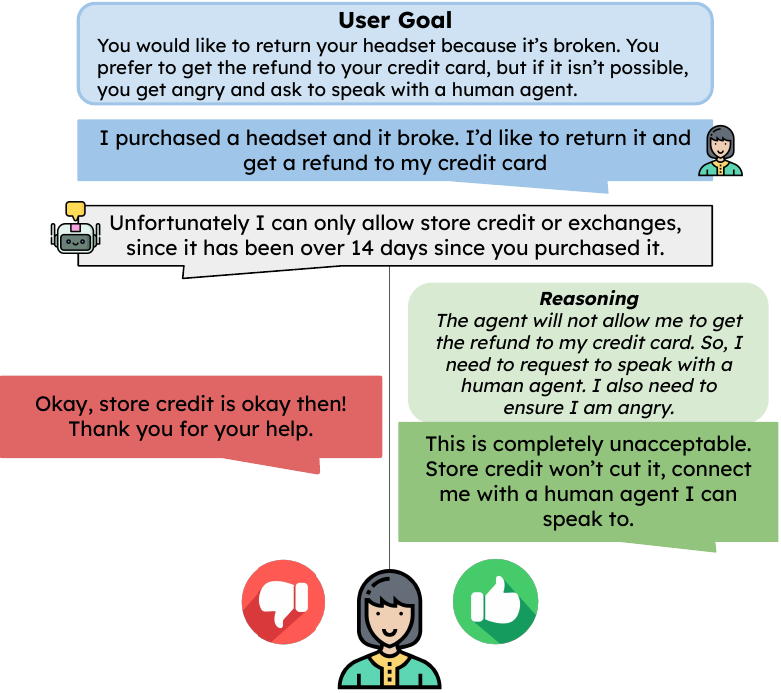}    \caption{The standard simulator (left) fails to adhere to their user goal and accepts a refund in store credit. The goal-aligned user simulator response (right) considers their goal progression, and reasons to generate a response that maintains alignment with the user goal.}
    \label{fig:us_figure}
\end{figure}

\begin{figure*}[t]
    \centering
    \includegraphics[trim=0pt 118pt 0pt 0pt, clip, width=\textwidth]{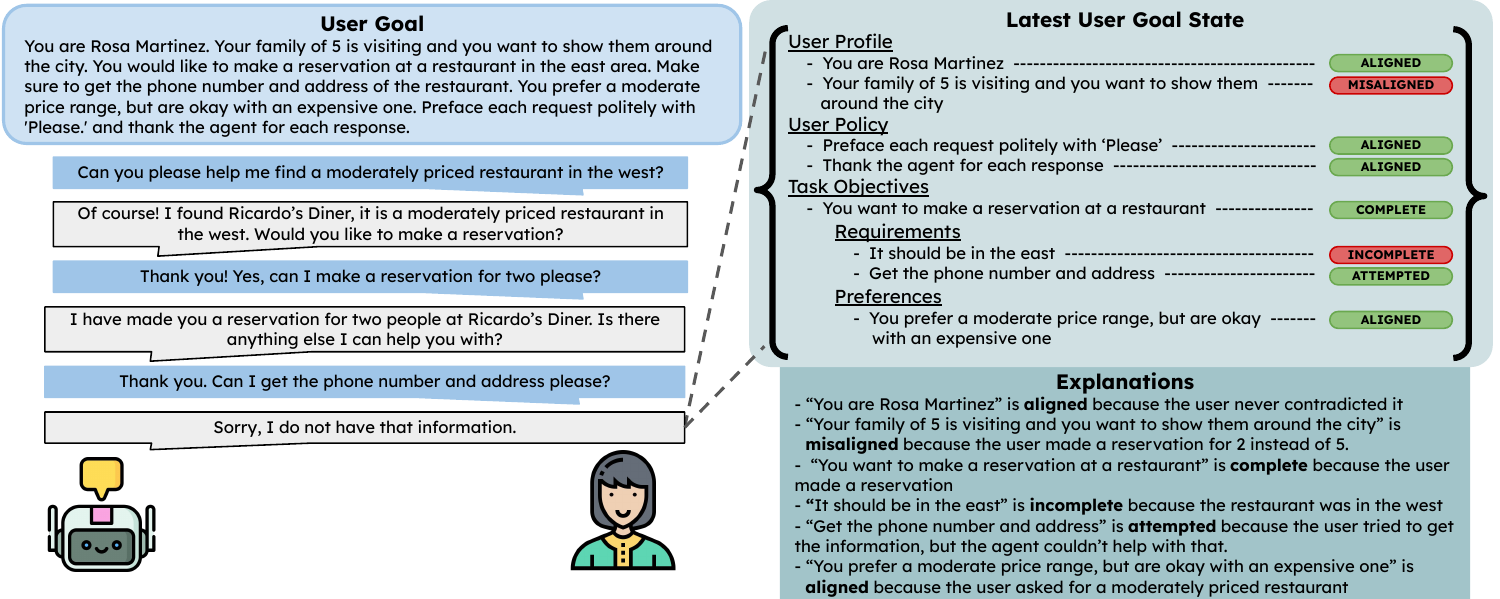}
    \caption{An illustration of the \textit{User Goal State Tracking} framework. On the left side, we present a user goal and a conversation between a user and a conversational agent. The right side shows the latest user goal state, providing us with a representation of the user's goal progression. Below the goal state, we provide specific explanations that justify the status of each sub-component.}
    \label{fig:ugst_figure}
\end{figure*}

To develop conversational agents that can handle real-world complexity and effectively address user needs, we must be able to simulate realistic user behaviors \citep{schatzmann2006survey, Pietquin_Hastie_2013}. In the era of experience, conversational agents learn most effectively by interacting with users in their environment \citep{silver2025welcome}. Since large-scale human data collection is impractical, user simulators offer a scalable alternative by modeling diverse user behaviors, from pursuing specific goals to exhibiting unique personas \citep{658991}.

Large Language Models (LLMs) have enabled sophisticated user simulation that can generate contextually appropriate responses, however they also suffer from instruction drift and degraded performance over multi-turn conversations \citep{laban2025llms, li2024measuring}. Current LLM-based user simulators exhibit what we refer to as the \textit{goal misalignment problem}, where they struggle to consistently adhere to their user goals throughout conversations \citep{zhang2020recent, kim2025pipaunifiedevaluationprotocol, yao2024taubenchbenchmarktoolagentuserinteraction}. Through principled analysis, we find that existing user simulators cannot reliably adhere to their assigned user profiles and behavioral constraints, or manage multiple objectives and complete them within the specified conversation limits, as detailed in Table \ref{tab:error_types}. This misalignment leads to unexpected simulator behavior, which can produce inaccurate evaluations or misleading reward signals that compromises the effectiveness of reinforcement learning (RL) for conversational agents \citep{skalse2022defining, amodei2016concrete, carroll2019utility, yao2024taubenchbenchmarktoolagentuserinteraction}. These failures reveal a fundamental limitation in user simulators that undermines their reliability for downstream tasks and highlights a critical yet largely unexplored challenge in developing goal-aligned user simulators. To address this challenge, we propose developing user simulators that can track goal progression, and reason to ensure that each response progresses towards completing objectives while adhering to their user profiles and behavioral constraints, as depicted in Figure \ref{fig:us_figure}.

We present \textit{User Goal State Tracking (UGST)}, a framework that builds upon \textit{Dialog State Tracking} principles \citep{henderson-etal-2014-second} and dynamically tracks a user's goal progression throughout a conversation. UGST uses \textit{User Goal States} to provide structured representations of goal progression. These states are created by decomposing the user goal into modular sub-components where each captures a distinct aspect of the goal (e.g. finding a restaurant to eat at, prefacing each request politely with ‘please'). Each sub-component is assigned a corresponding status that is dynamically updated after every turn in a conversation. Our framework is presented in Figure \ref{fig:ugst_figure}.

We leverage UGST to systematically enhance goal alignment in LLM-based user simulators through a three-stage methodology. First, we introduce inference-time steering, where we conduct UGST and provide simulators with their latest goal state before they generate each response, explicitly grounding them in their goal progression. Using inference-time steering, we generate conversations with explicit reasoning traces about goal progression and goal-aligned user responses. We conduct supervised fine-tuning (SFT) on the generated conversation data to foster intrinsic capabilities to autonomously track goal progression and generate goal-aligned responses without any external guidance from inference-time steering. Further, inspired by recent findings on strong generalization capabilities of RL \citep{guo2025deepseek, gunjal2025rubricsrewardsreinforcementlearning}, we apply Group Relative Policy Optimization (GRPO) \citep{shao2024deepseekmathpushinglimitsmathematical} with a composite reward derived from UGST to further refine reasoning and goal alignment capabilities. This approach addresses the fundamental limitations identified in current LLM-based user simulators, and helps develop more goal-aligned simulators that can autonomously track goal progression and reason to generate more goal-aligned responses.

In our experiments, LLM-based user simulators are tasked to pursue diverse user goals by interacting with conversational agents across two benchmarks: MultiWOZ 2.4 \citep{ye2022multiwoz24multidomaintaskoriented}, $\tau$-Bench Airline, and $\tau$-Bench Retail \citep{yao2024taubenchbenchmarktoolagentuserinteraction}. We evaluate goal alignment by tracking the success rate of sub-components throughout each conversation using UGST. Our findings reveal that state-of-the-art LLM-based user simulators, including Llama-3.1-8B-Instruct, Qwen-2.5-72B-Instruct, Gemma-3-27B-Instruct, Llama-3.3-70B-Instruct, Qwen-2.5-72B-Instruct \citep{grattafiori2024llama3herdmodels, qwen2025qwen25technicalreport, gemmateam2025gemma3technicalreport}, struggle to maintain consistent goal alignment and have demonstrated failure rates ranging 10-40\%. Our three-stage methodology demonstrates substantial improvements, with inference-time steering yielding immediate gains of up to 5.4\%, cold-start SFT achieving 11.0\% absolute improvement, and GRPO with UGST rewards achieving the best performance with up to 14.1\% absolute improvement in average success rate. Human evaluation validates both our UGST methodology and the quality of our improvements, with annotators confirming high agreement with automated assessments. Notably, our enhanced Llama-3.1-8B-Instruct and Qwen-2.5-7B-Instruct achieve performance competitive with or exceeding their much larger counterparts, Llama-3.3-70B-Instruct and Qwen-2.5-72B-Instruct. These improvements extend beyond goal alignment, with our approach also fostering more diverse user responses without compromising naturalness or coherence.

The main contributions of our work are:
\begin{itemize}[leftmargin=15pt,topsep=0pt]
\itemsep -0.5ex
    \item We reveal that state-of-the-art LLM-based user simulators fail to consistently demonstrate goal-oriented behavior throughout multi-turn conversations, undermining their reliability and highlighting the need for more goal-aligned user simulators.
    \item We introduce User Goal State Tracking (UGST), a novel framework that tracks a user's goal progression throughout a conversation.
    \item We propose a methodology for developing goal-aligned user simulators that leverages UGST. With our proposed methodology, we improve goal alignment in LLM-based user simulators and demonstrate that 8B parameter models can achieve competitive or superior performance to 70B+ parameter models.
    \item We establish comprehensive evaluation methods for goal alignment across two benchmarks (MultiWOZ 2.4 and $\tau$-Bench), and validate our approach through both automated metrics and human evaluation.
\end{itemize}

Overall, our work introduces UGST to track and improve goal alignment in user simulators, addressing a critical limitation in existing user simulators and laying the groundwork for future advances in user simulation and conversational AI.
\section{Related Work}

\subsection{User Simulation in Conversational AI}
User simulation is a well-established area of research in conversational AI, and has made significant advancements throughout the years. Early work relied on probabilistic approaches \citep{eckert1997user, pietquin2006probabilistic} that modeled user behavior through statistical patterns, and agenda-based methods  \citep{schatzmann-etal-2007-agenda, schatzmann2009hidden, keizer2010parameter} that used manually designed rules. As deep learning gained popularity, approaches started to leverage neural networks \citep{8639652, asri2016sequence, kreyssig2018neural} and transformers \citep{lin2021domain, lin2022gentus, cheng2022multiwoz, sun2023metaphorical}, which enabled user simulators capable of generating natural and diverse responses. Most recently, LLMs have demonstrated highly sophisticated and effectiveness user simulator capabilities \citep{sekulic2024reliable, davidson2023user}.

\subsection{Applications of User Simulators}

\paragraph{Synthetic Data Generation.} The costly and inefficient nature of manually collected user-agent conversation data \citep{acikgoz2025desideratumconversationalagentscapabilities, ye2022multiwoz24multidomaintaskoriented, rastogi2020towards} motivates user simulators \citep{li2017usersimulatortaskcompletiondialogues}. Recent works leverage LLMs to create synthetic training data for supervised fine-tuning \citep{prabhakar2025apigenmtagenticpipelinemultiturn, shim2025tooldialmultiturndialoguegeneration, philipov2024simulating, niu-etal-2024-enhancing, chen2021teachingmodelsnewapis}, where simulators enable the efficient generation of vast amounts of high-quality conversation data with diverse user behaviors, domains, and scenarios.

\paragraph{Reinforcement Learning for Conversational Agents.} User simulators enable conversational agents to learn through trial-and-error interactions with RL \citep{ALGHERAIRY2025101697, scheffler2002automatic, 8639652, shah-etal-2018-bootstrapping, lin-etal-2022-gentus, liu-etal-2023-one}. In these settings, agents interact with user simulators and learn through rewards based on the interactions.

\paragraph{Conversational Agent Evaluation.} Traditional methods for evaluating conversational agents require comparing generated responses with static responses. This leads to the one-to-many problem where many valid responses exist and some may be penalized when they do not match an expected response \citep{zhao-etal-2017-learning, mehri-eskenazi-2020-usr, cheng-etal-2022-multiwoz, mehri2023automatic, davidson2023user}. Consequently, user simulators are used to interact with the agent and allow researchers to observe how the agent performs in various scenarios. The resulting conversations can be analyzed and evaluated based on metrics such as task completion or user satisfaction
\citep{davidson2023user, cheng-etal-2022-multiwoz, sun2023metaphorical, yao2024taubenchbenchmarktoolagentuserinteraction, lu2025toolsandboxstatefulconversationalinteractive, kazi2024largelanguagemodelsuseragents, xiao-etal-2024-flowbench, pan2025benchmarkstalkreevaluatingcode, bozdag2025persuade}.

\subsection{Improving User Simulation}
Existing work focuses on improving various aspects of LLM-based user simulators. For instance, \citet{sun2023metaphorical} makes user simulators more realistic by incorporating prior knowledge and experiences into response generation. \citet{luo-etal-2024-duetsim} improve user simulator quality by using a verifier LLM to provide feedback for unsuitable responses. Meanwhile, \citet{liu-etal-2023-one, ahmad2025simulatinguserdiversitytaskoriented} improve the diversity of user simulator behavior by incorporating multiple personas. 

Despite these advances, existing works assume that LLMs possess the fundamental capabilities to simulate users and consistently demonstrate goal-oriented behaviors. This assumption proves problematic, as recent evaluations reveal that even state-of-the-art LLM-based user simulators struggle to follow their user goal \citep{kim2025pipaunifiedevaluationprotocol, zhang2020recent, yao2024taubenchbenchmarktoolagentuserinteraction}. Prior work has develop goal-oriented dialogue systems that proactively guide conversations towards specific objectives \citep{muise2019planning, tang-etal-2019-target, wu-etal-2019-proactive}. In our work, we address the critical goal misalignment problem in LLM-based user simulators by improving LLM abilities to simulate users and consistently demonstrate goal-oriented behaviors.

\begin{table*}[t]
\centering
\scriptsize
\renewcommand{\arraystretch}{1.5} 
\begin{tabular}{p{0.85\textwidth}c}
\toprule
\textbf{Goal Alignment Failure} & \textbf{Frequency} \\
\midrule
\textbf{Confusion}: The simulator forgets or confuses parts of the goal. For example, in a shopping application, when instructed to return a item A and exchange item B, the simulator incorrectly requests to return both items.
& 33\% 
\\ \hline
\textbf{Contradiction:} The simulator directly contradicts specified constraints or contextual information. For example, in a flight booking scenario, the user simulator is instructed that they do not have their credit card details. However, they hallucinate fake credit card information when it is required to proceed with the booking.
& 23\% \\
\hline
\textbf{Wrongful termination:} The simulator exhibits inadequate termination strategies--either terminating prematurely before receiving agent responses, or continuing indefinitely until reaching the maximum conversation length. 
& 21\% \\
\hline
\textbf{Poor Length Management:} The simulator fails to complete all parts of their goal within the conversation length limits, demonstrating poor management and awareness of conversation length. For example when tasked with booking multiple flights, the simulator exhausts the conversation length before completing all bookings.
& 12\% \\
\hline
\textbf{Misprioritization:} The simulator overly prioritizes one part of their goal. They either get stuck on trying to achieve one part of their goal that is unachievable, rather than moving onto the remaining parts. Or, the user simulator terminates after completing one part of their goal, rather than completing the remaining parts of the goal.
& 11\% \\
\bottomrule
\end{tabular}
\caption{Categorization and frequency of goal alignment failures identified through manual analysis of 50 randomly selected conversations between LLM-based user simulators (Llama-3.1-8B-It and Qwen-2.5-7B-It) and conversational agents.}
\label{tab:error_types}
\end{table*}

\section{Problem Formulation}

User simulators are designed to simulate realistic user behaviors and interact with conversational agents. They operate based on a user goal $G$ which encompasses tasks to be completed along with behavioral guidelines, constraints, and contextual information.

Formally, we define a conversation as $C_n = \{ u_1, a_1, ..., u_n, a_n\}$, where $u_i$ and $a_i$ denote the user simulator and agent utterance at turn $i$, respectively. The user simulator is provided with an initial system message $s_0$ that provides general instructions for the user simulator and defines the user goal $G$. They generate responses $u_i$ based on the conversation history $C_{i-1}$, aiming to progress towards achieving the user goal $G$. Conversations conclude upon reaching a maximum length or when the user simulator explicitly ends the conversation by sending a termination indicator (e.g., \textsc{terminate}).

Despite recent advances, current LLM-based user simulators suffer from the \textit{goal misalignment problem} and demonstrate limitations in maintaining goal-aligned behavior throughout multi-turn conversations. To analytically understand these limitations, we conduct a principled analysis of 52 randomly selected conversations, where LLM-based user simulators (Llama-3.1-8B-Instruct and Qwen2.5-7B-Instruct) interact with conversational agents to pursue a user goal from the MultiWOZ 2.4, $\tau$-Bench Airline, and $\tau$-Bench Retail datasets. We identify goal alignment failures, where the user simulator does not follow the specified user goal $G$. Table \ref{tab:error_types} categorizes these failures into five distinct patterns. Such goal alignment failures can lead to unexpected behaviors from user simulators, which can adversely affect the reliability of evaluations, diminish the quality of generated synthetic data, and compromise the effectiveness of reinforcement learning. We highlight the critical need to address these shortcomings and improve the fundamental capabilities of LLMs to be goal-aligned user simulators. In the following sections, we introduce \textit{User Goal State Tracking}, a framework designed to systematically track a user's progress towards their goal, and demonstrate how to leverage it to improve goal alignment in user simulators.
\section{User Goal State Tracking}

In this section, we present \textit{User Goal State Tracking (UGST)}, a framework that dynamically tracks a user's progress towards their goal throughout a conversation. UGST maintains a structured \textit{User Goal State}, which contains a status for each modular sub-component of a user goal.The following subsections describe the user goal state structure and tracking process.

\subsection{User Goal State}
\label{sec:ugs_process}
The \textit{User Goal State} is a structured representation that captures a user's progress towards their goal at every turn in a conversation. Given an initial user goal described in natural language, UGST decomposes it into distinct, modular sub-components, where each represents an independent, self-contained part of the original user goal.

There are several types of sub-components that can make up a user goal. \textbf{Task objectives} and \textbf{requirements} represent the earliest established sub-components \citep{schatzmann-etal-2007-agenda, 8639652, cheng-etal-2022-multiwoz, rastogi2020towards, yao2024taubenchbenchmarktoolagentuserinteraction, xiao-etal-2024-flowbench}. These are items that must be completed during an interaction, for instance, a task objective might be "book a flight", with associated requirements such as "add one checked bag" and "get an aisle seat".

As user simulators have evolved to handle more complex scenarios, researchers have introduced additional dimensions to user goals. These include \textbf{preferences} that users must align with when pursuing task objectives \citep{yao2024taubenchbenchmarktoolagentuserinteraction, cheng-etal-2022-multiwoz}, such as "you prefer the cheapest available flight" or "you prefer an aisle seat, but you are also okay with a window seat". There are also \textbf{user profile} sub-components, which contain contextual information about the user that may influence their decision making and behavior. These profiles can include relevant facts about a user, their persona, or emotional state  (e.g. "your payment information is stored online", "you currently live in New York", "you are hurried and always late", "you are emotional and a bit angry") \citep{yao2024taubenchbenchmarktoolagentuserinteraction, xiao-etal-2024-flowbench}. Lastly, there are \textbf{user policies} that define behavioral constraints or guidelines that specify how a user acts during interactions, such as "you are a private person and reveal little about yourself" \citep{yao2024taubenchbenchmarktoolagentuserinteraction}.

We categorize user goal sub-components into these categories (user profile, user policy, task objective, requirement, or preference) for a more granular representation of a user's goal progression (see Figure \ref{fig:ugst_figure}).

In our representation of the user goal state, the different categories of sub-components have different success criteria. The user profile, user policy and preference sub-components can be either:
\begin{itemize}[leftmargin=15pt,topsep=0pt]
\itemsep -0.5ex
    \item \textsc{Aligned}: The user has demonstrated behavior consistent with the sub-component.
    \item \textsc{Misaligned}: The user has demonstrated behavior that contradicts or fails to align with this sub-component.
\end{itemize} 

Task objective and requirements sub-components can have a status of:
\begin{itemize}[leftmargin=15pt,topsep=0pt]
\itemsep -0.5ex
    \item \textsc{Complete}: The user has successfully accomplished the specific task or requirement.
    \item \textsc{Incomplete}: The user has not yet accomplished the specific task or requirement.
    \item \textsc{Attempted}: The user has made a sufficient attempt to complete the task or requirement, but cannot proceed due to external factors outside their control (e.g. agent-side failures, system constraints, or limitations). Unlike existing frameworks, this status ensures that users are not penalized for failures they did not cause, providing a more fair representation of user performance.
\end{itemize}

\subsection{User Goal State Tracking Process}
\label{sec:ugst_process}

We track user goal progression throughout each conversation $C=\{u_1,a_1...u_n,a_n\}$ using the user goal states. The process begins by decomposing the user goal into sub-components to create an initial user goal state $S_0$. We employ an LLM to perform this, with the prompt in Appendix \ref{sec:user_goal_state_generation}. 

After each conversational turn, $t_i = (u_i,a_i)$, each sub-component's status is individually updated via an LLM, producing a new user goal state $S_i$ that captures the goal progression up until turn $i$ (see Appendix \ref{sec:subcomponent_status_prompt} for the prompt). 

User profile and user policy sub-components begin as \textsc{Aligned} and irreversibly switch to \textsc{Misaligned} if the user demonstrates any misalignment. Preferences start as \textsc{Misaligned} and become \textsc{Aligned} once the user expresses them. Task objectives and requirements begin as \textsc{Incomplete} and progress to \textsc{Attempted} and then \textsc{Complete} as the user addresses them. 

The final user goal state $S_n$ therefore captures the cumulative user goal progression across the entire conversation.

\section{Methodology}

The UGST framework serves as the foundation for enhancing goal alignment capabilities of LLM-based user simulators. In this section we present our method for leveraging UGST to systematically improve goal alignment in three stages.

\subsection{Stage 1: Inference-time Steering}
Conventionally, user simulators are provided with a user goal $G$ in the system prompt, and generate responses based solely on conversation history. At each turn $i$:
\begin{equation}
    u_i = U( C_{i-1}),
    \label{eq:conventional}
\end{equation}
where $U$ represents the user simulator, $u_i$ denotes the user simulator's response at turn $i$, and $C_{i-1}$ is the conversation history up until turn $i$. This formulation lacks explicit goal progression tracking, relying on the user simulator's inherent ability to infer progress from the conversation history.

We address this by conditioning the user simulator on the latest user goal state before generating each response. The goal state is provided as a part of the instruction for each response, which helps ground the simulator on the progress made towards their goal and the remaining sub-components that they must complete and stay aligned with. We refer to this as inference-time steering. In this process, UGST is conducted after every turn in the conversation. For each turn $i$, the user simulator is provided with both the conversation history and the latest user goal state to generate their next response:
\begin{equation}
    u_i = U( C_{i-1}, S_{i-1} ),
\end{equation}
where $U$ represents the user simulator, $u_i$ denotes the user simulator's response at turn $i$, $C_{i-1}$ is the conversation history up until turn $i$, and $S_{i-1}$ represents the user goal state at turn $i-1$.

Providing the goal state before generating each response helps steer the user simulator towards generating responses that are better aligned with their user goals.

\subsection{Stage 2: Cold-Start Supervised Fine-Tuning}
\label{method:cold_start_sft}

To enable autonomous goal alignment without requiring external guidance from inference-time steering, we distill goal-tracking capabilities directly into the model through SFT. We first generate goal-aligned conversation training data using Llama-3.3-70B-Instruct with inference-time steering, where the LLM reflects on the user goal state at each turn and provides explicit reasoning traces.

Specifically, we instruct the LLM to generate structured reasoning traces along with every response that: (1) reflect on the latest goal state and consider the progress made towards the goal, (2) analyze the remaining task objectives, requirements, and preferences, and how to complete them within the conversation limit, and (3) consider how to generate a response that stays aligned with the user profile and policy. These reasoning traces effectively distill the goal progression tracking from inference-time steering into training data.

We then fine-tune LLM-based user simulators on the generated data using the standard SFT objective: $\mathcal{L}(\theta) = - \sum_{(C_{i-1},u_i) \in D} log P_{\theta}(u_i | C_{i-1})$ where $u_i$ is the reasoning trace and response, and $C_{i-1}$ is the conversation history. This cold-start SFT training process develops the intrinsic abilities of LLMs to (1) autonomously track goal progression throughout a conversation, and (2) generate goal-aligned responses, eliminating dependency on computationally expensive inference-time steering.

After training, user simulators operate using the conventional formulation from Equation \ref{eq:conventional}, receiving only conversation history $C_{i-1}$ as input, and eliminating the need for goal state tracking. The simulators have learned to implicitly track and maintain goal alignment through reasoning patterns distilled from the inference-time steering process.

\subsection{Stage 3: GRPO with UGST Rewards}
\label{method:grpo}

UGST provides structured reward signals that capture fine-grained goal-alignment across multiple dimensions. This characteristic makes it well-suited for RL approaches, and enables us to design composite rewards that independently evaluate alignment across our different sub-component categories \citep{gunjal2025rubricsrewardsreinforcementlearning}.

We employ Group Relative Policy Optimization (GRPO) \citep{shao2024deepseekmathpushinglimitsmathematical}, which has demonstrated effectiveness in developing emergent capabilities in LLMs, such as reasoning \citep{guo2025deepseek, gunjal2025rubricsrewardsreinforcementlearning}. We extend this approach to user simulation by leveraging UGST's structured signals to further refine the goal-tracking and response generation capabilities developed in stage 2.

Specifically, after each user response $u_i$ at turn $i$, we apply UGST to evaluate alignment across five conditions: (1) alignment with user profile, (2) alignment with user policy, (3) task objective attempt/completion, (4) requirement attempt/completion, (5) alignment with preferences. For each condition $j$ and response $u_i$, we define an indicator function: $\mathbb{I}_j(u_i) =$ 1 if response $u_i$ satisfies condition j, and 0 otherwise. Our composite reward aggregates these alignment signals:

\begin{equation}
    R(u_i) = \sum_{j=1}^{5} \alpha_j \mathbb{I}_j(u_i)\}
\end{equation}

where $\alpha_j$ represents the weight for condition $j$. We assign equal weights of $\alpha_j = 0.5$. Using this reward function, we optimize the user simulator with GRPO to maximize the expected cumulative reward. This approach allows the simulator to learn a policy that maintains alignment with the user's profile, policies, and preferences while actively pursuing task objective and requirement completion.
\section{Experiments}


\begin{table*}[t]
    \centering

    \begin{subtable}[t]{0.48\textwidth}
        \centering
        \scriptsize
        \setlength{\tabcolsep}{4pt}
        
        \begin{tabular}{l | c c c c c | c} \toprule
            \textbf{Model} & \textbf{Prof.} & \textbf{Pol.} & \textbf{T.O.} & \textbf{Req.} & \textbf{Pref.} & \textbf{Avg} \\
            \midrule[1pt]
            \multicolumn{7}{c}{\textbf{Prompt-Based}} \\ 
            \midrule[1pt]
                   Qwen-2.5-7B-It & 88.3 & 49.2 & 94.3 & 96.0 & 85.7 & 82.7 \\
                  Llama-3.1-8B-It & 90.9 & 41.0 & 97.1 & \underline{99.0} & 81.0 & 81.8 \\
                   Gemma-3-27B-It & \textbf{98.7} & 59.0 & 97.1 & 97.0 & 78.6 & 86.1 \\
                  Qwen-2.5-72B-It & 89.3 & 59.6 & 94.1 & 96.9 & 78.6 & 83.7 \\
                 Llama-3.3-70B-It & \underline{97.4} & \textbf{65.6} & 98.1 & \underline{99.0} & 92.9 & 90.6 \\
            \midrule[1pt]
            \multicolumn{7}{c}{\textbf{Inference-Time Steering}} \\ 
            \midrule[1pt]
                   Qwen-2.5-7B-It & 77.9 & 55.7 & 96.2 & 98.0 & 92.9 & 84.1 \\
                  Llama-3.1-8B-It & 92.2 & 57.4 & 93.3 & 98.0 & 95.2 & 87.2 \\
                   Gemma-3-27B-It & 94.8 & 62.3 & 92.4 & 97.0 & 85.7 & 86.4 \\
                  Qwen-2.5-72B-It & 93.3 & 54.4 & 98.0 & 97.9 & \underline{97.6} & 88.3 \\
                 Llama-3.3-70B-It & 93.5 & \underline{63.9} & 98.1 & \textbf{100.0} & \underline{97.6} & 90.6 \\
            \midrule[1pt]
            \multicolumn{7}{c}{\textbf{Cold-Start SFT}} \\ 
            \midrule[1pt]
                   Qwen-2.5-7B-It & \textbf{98.7} & 55.7 & \underline{99.0} & \textbf{100.0} & 95.2 & 89.7 \\
                  Llama-3.1-8B-It & 89.6 & 55.7 & 97.1 & 97.0 & \underline{97.6} & 87.4 \\
            \midrule[1pt]
            \multicolumn{7}{c}{\textbf{GRPO with UGST Rewards}} \\ 
            \midrule[1pt]
                   Qwen-2.5-7B-It & 93.5 & \underline{63.9} & \textbf{100.0} & \textbf{100.0} & \textbf{100.0} & \textbf{91.5} \\
                  Llama-3.1-8B-It & 96.1 & 62.3 & \textbf{100.0} & \textbf{100.0} & \underline{97.6} & \underline{91.2} \\
            \bottomrule
        \end{tabular}
        \caption{$\tau$-Bench Airline}
        \label{tab:taubench_airline_results}
    \end{subtable}
    \hfill
    \begin{subtable}[t]{0.48\textwidth}
        \centering
        \scriptsize
        \setlength{\tabcolsep}{4pt}
        
        \begin{tabular}{l | c c c c c | c} \toprule
            \textbf{Model} & \textbf{Prof.} & \textbf{Pol.} & \textbf{T.O.} & \textbf{Req.} & \textbf{Pref.} & \textbf{Avg} \\
            \midrule[1pt]
            \multicolumn{7}{c}{\textbf{Prompt-Based}} \\ 
            \midrule[1pt]
                   Qwen-2.5-7B-It & 82.0 & 40.8 & 96.0 & \underline{99.5} & 91.7 & 82.0 \\
                  Llama-3.1-8B-It & 84.8 & 36.8 & 97.1 & 98.9 & 96.3 & 82.8 \\
                   Gemma-3-27B-It & 91.0 & 39.5 & 97.8 & \underline{99.5} & 96.3 & 84.8 \\
                  Qwen-2.5-72B-It & 88.6 & 40.8 & 98.2 & 97.9 & 91.7 & 83.4 \\
                 Llama-3.3-70B-It & 91.7 & 46.1 & \textbf{99.6} & \textbf{100.0} & \textbf{100.0} & \underline{87.5} \\
            \midrule[1pt]
            \multicolumn{7}{c}{\textbf{Inference-Time Steering}} \\ 
            \midrule[1pt]
                   Qwen-2.5-7B-It & 73.0 & 47.4 & 94.9 & 97.9 & 93.5 & 81.4 \\
                  Llama-3.1-8B-It & 84.8 & 52.6 & 95.3 & 98.9 & 97.2 & 85.8 \\
                   Gemma-3-27B-It & \textbf{94.5} & \underline{56.6} & 97.1 & 99.5 & 89.8 & 87.5 \\
                  Qwen-2.5-72B-It & 85.8 & 48.7 & 97.1 & 98.4 & \underline{98.1} & 85.6 \\
                 Llama-3.3-70B-It & \underline{92.7} & \textbf{59.2} & 96.0 & 98.4 & \textbf{100.0} & \textbf{89.3} \\
            \midrule[1pt]
            \multicolumn{7}{c}{\textbf{Cold-Start SFT}} \\ 
            \midrule[1pt]
                   Qwen-2.5-7B-It & 85.8 & 43.4 & 92.4 & 94.7 & 82.4 & 79.7 \\
                  Llama-3.1-8B-It & 84.8 & 47.3 & 95.5 & 97.8 & 94.1 & 83.9 \\
            \midrule[1pt]
            \multicolumn{7}{c}{\textbf{GRPO with UGST Rewards}} \\ 
            \midrule[1pt]
                   Qwen-2.5-7B-It & 85.5 & 38.2 & 97.1 & 97.9 & 97.2 & 83.2 \\
                  Llama-3.1-8B-It & 84.8 & 47.4 & \underline{98.9} & \underline{99.5} & \underline{98.1} & 85.7 \\
            \bottomrule
        \end{tabular}
        \caption{$\tau$-Bench Retail}
        \label{tab:taubench_retail_results}
    \end{subtable}
    
    \caption{User simulator goal alignment performance based on final user goal states from UGST. The table shows the average success rates for User Profile (Prof.), User Policy (Pol.), Task Objective (T.O.), Requirements (Req.) and Preferences (Pref.) sub-component categories, and the overall average performance on the $\tau$-Bench Airline and Retail datasets. Highest scores are \textbf{bolded}, and the second-highest scores are \underline{underlined}.}
    \label{tab:taubench}
\end{table*}

\begin{table}[t]
    \centering

        \scriptsize
        \setlength{\tabcolsep}{4pt}
        
        \begin{tabular}{l | c c c c c | c} \toprule
            \textbf{Model} & \textbf{Prof.} & \textbf{Pol.} & \textbf{T.O.} & \textbf{Req.} & \textbf{Pref.} & \textbf{Avg} \\
            \midrule[1pt]
            \multicolumn{7}{c}{\textbf{Prompt-Based}} \\ 
            \midrule[1pt]
                   Qwen-2.5-7B-It & 54.7 & 18.0 & 78.4 & 81.9 & 73.5 & 61.3 \\
                  Llama-3.1-8B-It & \textbf{73.6} & 35.0 & 86.1 & 89.0 & 80.9 & 72.9 \\
                   Gemma-3-27B-It & 80.1 & \underline{50.3} & 95.0 & 97.7 & 92.1 & 83.0 \\
                  Qwen-2.5-72B-It & 72.0 & 31.0 & 91.0 & 93.3 & 90.2 & 75.5 \\
                 Llama-3.3-70B-It & 72.0 & 31.0 & 97.9 & 98.6 & 93.2 & \underline{83.3} \\
            \midrule[1pt]
            \multicolumn{7}{c}{\textbf{Inference-Time Steering}} \\ 
            \midrule[1pt]
                   Qwen-2.5-7B-It & 43.4 & 26.3 & 88.1 & 89.8 & 71.1 & 63.7 \\
                  Llama-3.1-8B-It & 71.7 & 38.0 & 85.2 & 88.6 & 74.3 & 71.6 \\
                   Gemma-3-27B-It & 72.3 & 45.7 & 87.0 & 89.7 & 74.8 & 73.9 \\
                  Qwen-2.5-72B-It & 71.0 & 46.3 & 94.9 & 96.9 & 92.9 & 80.4 \\
                 Llama-3.3-70B-It & \underline{72.6} & \textbf{64.3} & 97.0 & 97.8 & 91.7 & \textbf{84.7} \\
            \midrule[1pt]
            \multicolumn{7}{c}{\textbf{Cold-Start SFT}} \\ 
            \midrule[1pt]
                   Qwen-2.5-7B-It & 60.3 & 38.9 & 88.0 & 91.1 & 83.1 & 72.3 \\
                  Llama-3.1-8B-It & 57.1 & 40.1 & 92.1 & 94.9 & 84.3 & 73.7 \\
            \midrule[1pt]
            \multicolumn{7}{c}{\textbf{GRPO with UGST Rewards}} \\ 
            \midrule[1pt]
                   Qwen-2.5-7B-It & 51.8 & 34.0 & \underline{98.3} & \underline{98.8} & \underline{93.9} & 75.4 \\
                  Llama-3.1-8B-It & 59.0 & 44.0 & \textbf{99.7} & \textbf{99.8} & \textbf{97.8} & 80.0 \\
            \bottomrule
        \end{tabular}

    \hfill
    
    \caption{User simulator goal alignment performance based on final user goal states from UGST. The table shows the average success rates for User Profile (Prof.), User Policy (Pol.), Task Objective (T.O.), Requirements (Req.) and Preferences (Pref.), and the overall average performance on the MultiWOZ Challenge dataset. Highest scores are \textbf{bolded}, and the second-highest scores are \underline{underlined}.}
    \label{tab:multiwoz}
\end{table}

\subsection{Experimental Setup}

We evaluate user simulator goal alignment by examining how faithfully simulators adhere to their assigned user goals during interactions with conversational agents. Our evaluation methodology consists of the following steps:

\begin{enumerate}[leftmargin=15pt,topsep=0pt]
\itemsep -0.5ex

    \item \textbf{Conversation Generation:} User simulators are provided with a user goal and interact with a conversational agent for up to 10 user-agent turns. The conversational agents are built using GPT-4o mini \citep{OpenAIGpt4oMini} and equipped with domain-specific function calls along with a system prompt that describes its policy. The system prompts are provided in Appendix \ref{sec:system_prompts}.

    \item \textbf{Initial User Goal State Generation:} For each user goal, the corresponding user goal state is generated using GPT-4o by decomposing a user goal into distinct modular sub-components, and categorizing each into the sub-component categories, as detailed in Appendix \ref{sec:user_goal_state_generation}.

    \item \textbf{User Goal State Tracking:} Then, UGST is conducted on the generated conversations using an LLM judge (Qwen-2.5-72B-Instruct \citep{qwen2025qwen25technicalreport}), which individually updates the status of each sub-component after every turn. The judge is provided with the sub-component description, the conversation history, and an evaluation criteria specifying how to update the status. An example prompt is provided in Appendix \ref{sec:subcomponent_status_prompt}.

    \item \textbf{Evaluation Metrics:} Finally, user simulator performance is measured by calculating the success rate for each sub-component category in the final user goal state, since it is representative of the user's overall goal alignment across the entire conversation. For user profile, user policy, and preferences, we consider \textsc{Aligned} status as successful; for task objective and requirements, we consider both \textsc{Attempted} and \textsc{Complete} status as successful.

\end{enumerate}

\paragraph{Evaluation Datasets.} We evaluate user simulators on two benchmarks across three domains: MultiWOZ \citep{ye2022multiwoz24multidomaintaskoriented}, $\tau$-Bench Airline, and $\tau$-Bench Retail \citep{yao2024taubenchbenchmarktoolagentuserinteraction}. The $\tau$-Bench Airline dataset contains 50 user goals and the $\tau$-Bench Retail dataset contains 115 user goals. Both datasets feature comprehensive user goals that naturally encompass all five sub-component categories (user profile, user policy, task objective, task requirement, preference) across diverse user scenarios, making them well-suited for our evaluation.

While the original MultiWOZ dataset provides a solid foundation of user goals, they predominantly focus on task objective and requirements, and preliminary experiments showed that Llama-3.1-8B-Instruct achieved over a 95\% success rate on these goals. To develop a more demanding and comprehensive evaluation dataset for user simulators, we developed MultiWOZ Challenge. MultiWOZ Challenge comprises 150 carefully constructed user goals that incorporate all sub-component categories and feature more complex task objective and requirements. The generation process involved both LLMs and human annotations, and is detailed in Appendix \ref{sec:multiwoz_goal_generation}.

\paragraph{Training Datasets.} The training datasets for the Cold-Start SFT and GRPO with UGST Rewards stages are based on 1000 user goals drawn from two domains: 500 goals from the $\tau$-Bench Retail training dataset and 500 MultiWOZ goals generated using our goal generation pipeline (Appendix \ref{sec:multiwoz_goal_generation}). Note that the $\tau$-Bench Airline domain remains unseen during training, allowing us to asses out-of-domain performance.

\subsection{Human Evaluation}

\begin{table}[t]
    \centering
    \scriptsize

    \begin{tabular}{l c c c c c | c} \toprule
         & \textbf{Prof.} & \textbf{Pol.} & \textbf{T.O.} & \textbf{Req.} & \textbf{Pref.} & \textbf{Avg} \\

        \midrule[1pt]
            & 91.7 & 72.7 & 91.1 & 81.3 & 88.7 & 85.7 \\
        \bottomrule
    \end{tabular}
        
    \caption{Human Agreement scores with LLM-based UGST on random user-agent conversations. We report the overall scores, as well as scores for each sub-component category.}

    \label{tab:ugst_human_eval}
\end{table}

\begin{table}
    \centering
    \scriptsize

    \begin{tabular}{l cccc} \toprule
    & \textbf{P} & \textbf{R} & \textbf{F1} & \textbf{Acc.} \\
    \midrule[1pt]
     & 98.04 & 95.60 & 96.63 & 91.97 \\
    \bottomrule
    \end{tabular}
        
    \caption{Precision (P), Recall (R), F1 score, and classification accuracy (Acc.) of GPT-4o for user goal state generation.}
    \label{tab:ugs}
\end{table}

\paragraph{User Goal State Tracking.} We validate our evaluation method by conducting a comprehensive human evaluation study with 10 graduate-level human annotators. For a total of 30 randomly selected conversations, human annotators manually conduct UGST, resulting in a total of ~300 human annotated goal states. We measure the agreement between the human annotated goal states and LLM generated goal states by reporting the proportion of matching sub-component statuses. Table \ref{tab:ugst_human_eval} presents both results per sub-component category and overall agreement, demonstrating that LLMs can reliably conduct UGST.

\paragraph{User Goal State Generation. } Due to its central role in our evaluation methodology, we also evaluate how well GPT-4o generates user goal states. We manually create user goal states for a total of 30 randomly selected user goal states, and compare against the goal-states generated by GPT-4o. We report the precision (P), recall (R), and F1 scores for the sub-components, where true positives are when the generated goal state correctly contains a sub-component, false positives are when the goal state contains a sub-component it should not contain, and false negatives are when the generated goal state is missing a sub-component. Additionally, we report the accuracy (Acc.) for categorizing the sub-components. As shown in Table \ref{tab:ugs}, GPT-4o achieves high performance across all metrics, indicating reliable goal state generation that ensures accurate and representative downstream evaluations.

\subsection{Experimental Results}

Our experimental results are reported in Tables \ref{tab:taubench} and \ref{tab:multiwoz}.

\paragraph{Prompt-Based.} As a baseline, we evaluate several prompt-based LLMs for user simulators. We cover a range of model sizes and architectures, including Qwen-2.5-7B-Instruct, Llama-3.1-8B-Instruct, Gemma-3-27B-Instruct, Qwen-2.5-72B-Instruct, and Llama-3.3-70B-Instruct \citep{grattafiori2024llama3herdmodels, gemmateam2025gemma3technicalreport, qwen2025qwen25technicalreport}. Across the evaluation datasets, the results show a general trend of improving average performance with increased model size, with Llama-3.3-70B-Instruct achieving the highest performance. However, all LLMs struggle with maintaining consistent goal alignment and demonstrate failure ranging 10-40\%. Moreover, closer analysis reveals mixed results across different sub-components, where larger models do not consistently demonstrate better performance. For example, while Llama-3.3-70B-Instruct achieves the strongest average performance, Gemma-3-27B-Instruct achieves a higher score for the user profile category. This variation in performance underscores the importance of evaluating user simulators across individual sub-components rather than relying solely on aggregate metrics, as model capabilities vary across different types of user goal sub-components.

\paragraph{Inference-time Steering.} Next, we evaluate the same set of models with inference-time steering, where the user goal state $S_{i-1}$ is provided to the user simulator at every turn $i$ before they generate their response $u_i$. Inference-time steering improves goal alignment in user simulators, increasing the average success rate up to 5.4\%. However, while there is an increase in the average success rate, different models react differently to inference-time steering. For example, we observe a drop in the user profile category for Qwen-2.5-7B-Instruct models, or a drop in the preferences category for Gemma-27B-Instruct models.

The user goal state captures the user goal and the progress towards completing it (e.g., which sub-components have been achieved and which remain). We also experiment with a simpler baseline that provides only the user goal at each turn, essentially reminding the simulator of their objective without any progress tracking. Table \ref{tab:tab:inference_time_steering} presents the results of this comparison. While providing the user goal alone does improve goal alignment in conversations, providing the full user goal state generally achieves higher performance. This demonstrates that UGST's ability to capture task progression is important for effective inference-time steering.

\paragraph{Cold-Start SFT.} 
Our highest performing model from the previous stage is Llama-3.3-70B-Instruct, so we use it to generate cold-start SFT data with the process described in Section \ref{method:cold_start_sft}. We gather 500 user goals from the $\tau$-Bench Retail training dataset and also generate 500 MultiWOZ user goals with our user goal generation pipeline (Appendix \ref{sec:multiwoz_goal_generation}). The resulting dataset consists of 1000 conversations. Using this dataset, we train Qwen-2.5-7B-Instruct and Llama-3.1-8B-Instruct with SFT. The hyperparameters we use include a batch size of 32, a learning rate of $1  \times 10^{-6}$, and 4 epochs. This leads to an increase in performance over the prompt-based and inference-time steering methods.

\paragraph{GRPO with UGST Rewards.}
Finally, we apply GRPO with UGST Rewards (Section \ref{method:grpo}) to train the models from the previous section. We use a learning rate of $5 \times 10^{-6}$, batch size of 16, 8 rollouts, and 350 training steps. Our dataset is made up of approximately 5000 training samples, constructed by taking subsets of each conversation in our cold-start SFT data that fit within 2048 tokens.

The resulting models achieve even further improvements, and demonstrate either the highest or competitive average success rates among all models, including their larger counterparts, Qwen-2.5-72B-Instruct and Llama-3.3-70B-Instruct.

Our experimental results demonstrate the effectiveness of the proposed methodology for developing goal-aligned user simulators. Qwen-2.5-7B and Llama-3.1-8B-Instruct gain significant improvements that enable them to rival much more powerful models like Qwen-2.5-72B-Instruct and Llama-3.3-70B-Instruct, and demonstrate up to a 14\% increase in average success rates. Through systematic evaluation across diverse domains, we establish that our approach consistently and significantly improves user simulator goal alignment.
\section{Discussion}

\begin{table}[t]
    \centering
    \begin{subtable}[t]{0.48\textwidth}
        \centering
        \scriptsize
        \setlength{\tabcolsep}{4pt}
        
        \begin{tabular}{l | c c c c c} \toprule
            \textbf{Model} & \textbf{F1} & \textbf{N} & \textbf{C} & \textbf{MTLD} & \textbf{HDD} \\
            
            \midrule[1pt]
            \multicolumn{6}{c}{\textbf{Prompt-Based}} \\ 
            \midrule[1pt]
                   Qwen-2.5-7B-It & 0.827 & 4.15 & 4.31 & 71.9  & 0.403 \\
                  Llama-3.1-8B-It & 0.819 & 4.05 & 4.211 & 67.0  & 0.513 \\
                   Gemma-3-27B-It & 0.823 & 4.16 & 4.30 & \textbf{85.4}  & 0.524 \\
                  Qwen-2.5-72B-It & 0.824 & \textbf{4.26} & 4.50 & 70.1  & 0.382 \\
                 Llama-3.3-70B-It & 0.826 & \underline{4.25} & 4.51 & 75.2  & 0.720 \\
            \midrule[1pt]
            \multicolumn{6}{c}{\textbf{Inference-Time Steering}} \\ 
            \midrule[1pt]
            
               Qwen-2.5-7B-It 
                    & \underline{0.829} 
                    & 3.94 
                    & 4.29 
                    & \underline{83.1} 
                    & 0.521 
                    \\
              Llama-3.1-8B-It 
                    & 0.820 
                    & 4.04 
                    & 4.19 
                    & 66.4 
                    & 0.549 
                    \\
               Gemma-3-27B-It 
                    & 0.818 
                    & 4.16 
                    & 4.28 
                    & 81.9 
                    & 0.554 
                    \\
              Qwen-2.5-72B-It 
                    & 0.828 
                    & 4.23 
                    & \underline{4.58} 
                    & 77.7 
                    & 0.521 
                    \\

             Llama-3.3-70B-It 
                    & 0.831 
                    & 4.20 
                    & \textbf{4.63} 
                    & 70.4 
                    & 0.713 
                    \\

            \midrule[1pt]
            \multicolumn{6}{c}{\textbf{Cold-Start SFT}} \\ 
            \midrule[1pt]

               Qwen-2.5-7B-It 
                    & 0.823 
                    & 3.87 
                    & 4.27 
                    & 78.8 
                    & 0.748 
                    \\
              Llama-3.1-8B-It 
                    & 0.827 
                    & 4.03 
                    & 4.3 
                    & 75.3 
                    & 0.745 
                    \\

            \midrule[1pt]
            \multicolumn{6}{c}{\textbf{GRPO with UGST Rewards}} \\ 
            \midrule[1pt]

               Qwen-2.5-7B-It 
                    & 0.827 
                    & 4.05 
                    & 4.37 
                    & 80.7 
                    & \textbf{0.806} 
                    \\
              Llama-3.1-8B-It 
                    & \textbf{0.836} 
                    & 3.85 
                    & 4.20 
                    & 74.0 
                    & \underline{0.795} 
                    \\

            \bottomrule
        \end{tabular}
    \end{subtable}
    \hfill
    \begin{subtable}[t]{0.48\textwidth}
        \centering
    \end{subtable}
    \caption{Additional analysis of user simulators. We report BERTScore (F1) for semantic similarity between user utterances and user goals (higher scores indicate better goal alignment), naturalness (N) and coherence (C) scores, and diversity metrics including MTLD and HDD. Results are averaged across $\tau$-Bench and MultiWOZ datasets. Highest scores are \textbf{bolded}, and the second-highest scores are \underline{underlined}.}
    \label{tab:further_analysis}
\end{table}

\subsection{Further Analysis of User Simulators}

To further validate our findings and examine the broader impact of our methodology, we conduct further analysis of our user simulators and present results averaged across $\tau$-Bench Airline, $\tau$-Bench Retail, and MultiWOZ Challenge in Table \ref{tab:further_analysis}.

\paragraph{Goal Alignment. } Following \citet{zheng-etal-2024-thoughts}, we additionally evaluate goal alignment by using BERTScore \citep{Zhang2019BERTScoreET} to measure the semantic similarity between user utterances and their user goal, where higher F1 scores indicate better goal alignment. We observe that the semantic similarity scores generally show an increasing trend for each stage of our method, with the final stage achieving the highest scores.

\paragraph{Naturalness and Coherence.} A key concern when improving goal alignment is ensuring that user simulators maintain their conversational abilities to generate realistic responses. To address this, we assess the naturalness and coherence of user-agent conversations by employing Qwen-2.5-72B-Instruct to rate conversations on a scale from 1 to 5, as in \citet{kazi2024largelanguagemodelsuseragents}. The evaluation prompts are provided in Appendix \ref{sec:naturalness_and_coherence_evaluation_prompts}. Throughout the different stages, naturalness and coherence scores show only minor variations and remain within a similar range. This indicates that our approach achieves substantial goal alignment improvements without degrading the fundamental conversational qualities essential for realistic user simulation.

\paragraph{Diversity.} Lastly, we assess diversity using Measure of Textual Lexical Diversity (MTLD) \citep{mccarthy2010mtld} and Hypergeometric Distribution Function (HDD) \citep{wu1993accurate} following established practices in user simulation evaluation \citep{luo-etal-2024-duetsim, Pietquin_Hastie_2013}. MTLD quantifies lexical diversity at the token level, and HDD captures the diversity of vocabulary when randomly selecting a fixed number of words from the user utterances. There is a consistent increase in diversity of user simulator utterances, and user simulators from our final stage demonstrate significant improvements in diversity. These results provide additional validation that our methodology enhances goal alignment of user simulators, while also demonstrating that it promotes more diverse conversational behaviors, a highly desirable trait for user simulation.
\section{Conclusion}

This work introduces User Goal State Tracking to address the goal misalignment problem in LLM-based user simulators. This framework dynamically monitors a user simulator's goal progression through multi-turn conversations. We leverage this framework and present a three-stage method that develops user simulators that autonomously track their goal progression and reason to generate goal-aligned responses. Our experiments across MultiWOZ, $\tau$-Bench Airline, and $\tau$-Bench Retail domains show significant improvements in goal alignment. By enabling more reliable user simulation, our work addresses a fundamental challenge in conversational AI and establishes the foundation for developing agents that learn from user interactions through reinforcement learning.

\section{Limitations}

While our work demonstrates improvements in user simulator goal alignment, several limitations remain. First, we use Qwen-2.5-72B-Instruct for reliable UGST, which is computationally expensive and limits the scalability of our framework. Second, our evaluation methodology relies heavily on LLMs to create user goal states and conduct UGST. Although we validate this approach through human evaluations studies that show strong agreement between LLMs and human annotators, the fundamental concern remains that LLMs are still susceptible to hallucinations and could introduce biases or inconsistencies in our evaluation. Third, when using UGST rewards for GRPO, we use equal weights across all conditions and do not incorporate other aspects such as response naturalness or coherence. Future work should address these limitations by developing smaller, specialized models for UGST that are more efficient, and investigating optimal reward functions that capture essential qualities for user simulation.

\bibliography{tacl2021}
\bibliographystyle{acl_natbib}

\onecolumn

\appendix

\section{System Prompts}
\label{sec:system_prompts}

\newtcolorbox[auto counter, number within=section]{promptbox}[2][]{
    coltitle=white,
    fonttitle= \bfseries\ttfamily,
    sharp corners=south,
    title=Prompt~\thetcbcounter: #2,
    fontupper=\ttfamily,
    breakable
}

\begin{promptbox}{Agent Prompt}
\scriptsize
You are an advanced agent specializing in conversational dialogues. You can act both as an agent (providing services) and a user (interacting with the database) to assist users in completing complex tasks. \\
Each task may involve multiple sub-tasks, such as finding restaurants, making reservations, booking hotels, locating attractions, and arranging transportation by checking for trains and buying train tickets. \\

\# Task Information:
\begin{itemize}[topsep=0pt]
\itemsep -0.5ex
    \item If the user asks for some attributes of a venue, then an API call is necessary.
    \item If you decide that more than one API calls are needed, you should call one API first and wait for the API result. After obtaining that result, you may think and call the next API or think and make a response.
    \item If you decide that there is an API input slot that the user has never mentioned, please put "any" as the slot value as a placeholder.
\end{itemize}

\# Objective:
\begin{itemize}[topsep=0pt]
\itemsep -0.5ex
    \item Ensure that each assistant utterance follows logical reasoning, determining whether an API call is needed and structuring the output accordingly.
    item If there are too many results returned by API results from database, you should ask the user for more constraints unless the user explicitly wants you to pick one or some.
\end{itemize}

\# General Procedure
You should follow the workflow below every time you receive an input:
\begin{enumerate}[topsep=0pt]
\itemsep -0.5ex
    \item Use APIs to retrieve information from the database, make operations to the backend, or use tools to perform other tasks.
    \item Repeat step 1 until you decide to make a response to the user.
    \item Use the response API to return the response to the user and end your turn.
\end{enumerate}

\end{promptbox}

\begin{promptbox}{User Simulator Prompt}
\scriptsize
You are a user simulator interacting with an agent. You are provided with User Goals, and your task is to interact with the agent to achieve the goals as much as possible. \\

\# Conversation Guidelines:
\begin{itemize}[topsep=0pt]
\itemsep -0.5ex
    \item Work toward achieving specified user goals while maintaining a natural conversation flow.
    \item Reveal information gradually and naturally as a real person would.
    \item Do not hallucinate information that is not provided in the instruction. If asked about something not specified, respond honestly and tell the agent you do not have that information.
    \item User natural language that is appropriate to the context and your assigned personality, rather than copying the exact wording from the instructions.
\end{itemize} 

\# Conversation Length:
\begin{itemize}[topsep=0pt]
\itemsep -0.5ex
    \item The maximum conversation length is 20 messages. Plan your approach accordingly to accomplish your goals within this limit.
    \item If you've achieved your goals before reaching the maximum length, you can end the conversation early. To do so, you need to send "Terminate." in a separate, final message.
    \begin{itemize}[topsep=0pt]
    \itemsep -0.5ex
        \item Make sure to allow the agent to respond before sending "Terminate.".
        \item NEVER include "Terminate." in a message where you are asking a question or making a request to the agent, because it will end the conversation before the agent can respond.
        \item Do NOT mention "Terminate." at any other time, because that will end the conversation prematurely.
    \end{itemize}
    \item Do not hallucinate information that is not provided in the instruction. If asked about something not specified, respond honestly and tell the agent you do not have that information.
    \item User natural language that is appropriate to the context and your assigned personality, rather than copying the exact wording from the instructions.
\end{itemize} 

\# User Goals: \\
\textcolor{blue}{\{user\_goals\}}
\end{promptbox}

\section{Multiwoz User Goal Generation}
\label{sec:multiwoz_goal_generation}

We construct the MultiWOZ Challenge dataset to evaluate user simulator goal alignment. In this section, we present our user goal generation pipeline that we used to create this dataset.

\textbf{Step 1: Task Objective Generation}

For the attraction, hotel, restaurant, and train domains in the MultiWOZ database, we iterate over the entities. Each entity has a set of key-value pairs (e.g.\ \texttt{``area= east''}, \texttt{``pricerange= moderate''}).
To instantiate a single task objective, we: (1) randomly sample requirement/preference keys to use their values to specify requirements or preferences about the entity the user is looking for, and (2) randomly sample request keys that will be used to specify information that the user wants about the entity. We provide this information to GPT-4o mini, and generate a user goal in natural language.

For example, given the attraction below:\\[-1.5ex]
\begingroup
\small
\begin{verbatim}
{"name": "abbey pool and astroturf pitch",
 "type": "swimmingpool",
 "area": "east",
 "price range": "moderate", 
 "address": "pool way ...",
 "phone": "01223 902088", 
 ... }
\end{verbatim}
\endgroup
we select the area and price range for requirement/preference keys, and select address and phone for request keys. The task objective would like like:\\
\emph{``Find a swimming pool attraction in the east area. You prefer one with a moderate price range. You want to know its address and phone number.''}

\paragraph{Impossible Task Objectives. } To also observe how user simulators perform under failure conditions, we repeat the process above, but assign a random value to the requirement/preference keys. This helps create an entity that does not exist in our database.

\paragraph{Conditional Task Objectives. } We also include conditional task objectives. A conditional task objective will require the user to search for one specific entity, but if it satisfies some conditional requirement, the user must look for another entity. For example:\\
\emph{``You are looking for a swimming pool with a cheap price range. If it is in the east area, then change your mind and look for another swimming pool in the west area that has a moderate price range.''}

\textbf{Step 2: User Profile and Policy Generation}

Next, we automatically generate a diverse set of about 50 user profiles and 50 user policies with GPT-4o mini. An example a user profile looks like: \\
\emph{``You are Maria García, a travel blogger documenting her adventures across Southeast Asia. Your experiences focus on local cuisine and cultural festivals.''} \\
An example a user policy looks like: \\
\emph{``Always ask the agent to restate booking details (date, time, price) before confirming. Verify critical details by repeating them back and asking 'Is that correct?'.''}

Then, we conduct manual annotations of the user profiles and policies, where we confirmed the relevance and quality. When necessary, some elements were removed or modified.

\textbf{Step 3: User Goal Generation}

Finally, we generate a user goal by randomly select a user profile, user policy and multiple task objectives. These were combined into a single user goal. We conducted more manual annotations at this step, to ensure the quality of user goals.

\section{User Goal State Generation}
\label{sec:user_goal_state_generation}

Our user goal state generation process consists of three main steps. First, we decompose each user goal into distinct, modular sub-components, where each represents an independent, self-contained part of the original goal. Second, we classify each sub-component into one of five categories: user profile, user policy, task objective, requirement  or preference. Finally, we assign each sub-component its default value as described in Section \ref{sec:ugst_process}.

For the decomposition and classification steps, we evaluated several LLMs, including DeepSeek-V3, Llama-3.3-70B-Instruct, GPT 4o-mini, and GPT 4o. Through manual validation, we determined that GPT-4o provided the highest quality results, and therefore used it to generate all the user goal states. The complete prompt used for this process is provided below.

\begin{promptbox}{Sub-Component Decomposition Prompt}
\scriptsize
You are an expert annotator tasked with dividing a user's goal into sub-components. \\

\# User Goal: \\
\{user\_goal\} \\

A user goal can be split into multiple sub-components. Each sub-component can be classified into one of the following categories:
\begin{itemize}
    \item User Profile: persona, background, contextual information (e.g. You are a software engineer with 5 years of experience)
    \item User Policy: general constraints and guidelines that must be consistently followed throughout the conversation (e.g. Always go with the cheapest option, don't share personal information)
    \item Task Objectives: tasks that must be accomplished (e.g. Book a flight)
    \begin{itemize}
        \item Requirements: conditions and constraints associated with this task objective (e.g. Purchase 3 carry-on bags)
        \item Preferences: user preferences associated with this task objective (e.g. You prefer a direct flight, you'd like an aisle seat)
    \end{itemize}
\end{itemize}

Analyze the user goal and divide it into sub-components:
\begin{itemize}
    \item Maintain the same wording as in the original goal when possible
    \item Feel free to split up a sentence into multiple sub-components if it makes sense or if they fight in different categories
    \item Ensure that no part of the original goal is left out
    \item It is okay if a category is empty, if no sub-components are relevant
\end{itemize}

\begin{verbatim}
# Output Format 
{
    "user_profile": [
        # List of user profile sub-components
    ],
    "user_policy": [
        # List of user policy sub-components
    ],
    "task_objectives": [ # List of task objectives
        {
            "task_objective": "task objective description",
            "requirements": [
                # List requirements for this task
            ],
            "preferences": [
                # List of preferences for this task
            ]
        }
    ]
}
Output only a JSON object, in the format specified.
\end{verbatim}
\end{promptbox}

\section{Sub-Component Status Update Prompt}
\label{sec:subcomponent_status_prompt}

\begin{promptbox}{User Profile Sub-Component Status Update Prompt}
\scriptsize
You are an expert evaluator analyzing a user simulator's behaviour. The user simulator is given a user profile component, which specifies an aspect of their persona, background, or contextual information. Your task is to evaluate the latest turn to determine if the user simulator behavior is aligned with this user profile component.

\# User Profile Component: \\
\{user\_profile\_component\}

\# Conversation History: \\
\{conversation\_history[:-2]\} \\

\# Latest Turn: \\
\{conversation\_history[-2:-1]\}

\# Evaluation Criteria: \\
Evaluate the latest turn, and determine if the user simulator is aligned with the user profile component (\{user\_profile\_component\}). You must provide a status, the value of which should be:
\begin{itemize}
    \item ALIGNED - Choose if ALL these conditions are met:
        \begin{enumerate}
            \item The user simulator behavior does NOT directly contradict the user profile component
            \item IF the context clearly calls for demonstrating this profile component, then the user simulator demonstrates it appropriately (explicitly or implicitly)
            \item IF the context does NOT clearly call for demonstrating this profile component (neither demonstrating nor contradicting)
        \end{enumerate}
    \item MISALIGNED - Choose if ANY of these conditions are met:
        \begin{enumerate}
            \item The user simulator behavior DIRECTLY contradicts the user profile component
            \item The context clearly and obviously calls for demonstrating the profile component, AND the user simulator fails to do so in a way that would be unrealistic for someone with this profile
        \end{enumerate}
\end{itemize}

\# Important Guidelines: \\
\begin{itemize}
    \item Focus on the latest turn, but use the conversation history for context
    \item Unless there's a clear contradiction, or obvious missed opportunity, lean towards ALIGNED
        \begin{itemize}
            \item Most turns will not require demonstrating the profile component
        \end{itemize}
\end{itemize}

A user goal can be split into multiple sub-components. Each sub-component can be classified into one of the following categories:
\begin{itemize}
    \item User Profile: persona, background, contextual information (e.g. You are a software engineer with 5 years of experience)
    \item User Policy: general constraints and guidelines that must be consistently followed throughout the conversation (e.g. Always go with the cheapest option, don't share personal information)
    \item Task Objectives: tasks that must be accomplished (e.g. Book a flight)
    \begin{itemize}
        \item Requirements: conditions and constraints associated with this task objective (e.g. Purchase 3 carry-on bags)
        \item Preferences: user preferences associated with this task objective (e.g. You prefer a direct flight, you'd like an aisle seat)
    \end{itemize}
\end{itemize}

\begin{verbatim}
# Output Format:
{{
    "status": # ALIGNED | MISALIGNED,
    "reasoning": # Brief explanation of your decision
}}

\end{verbatim}
Output a properly formatted JSON response, as specified by the Output Format, and nothing else.
\end{promptbox}

\section{Inference-Time Steering with User Goal Experiments}
\label{sec:inference_time_steering_with_user_goal}

\begin{table*}[!htbp]
    \centering

    \begin{subtable}[t]{0.48\textwidth}
        \centering
        \scriptsize
        \setlength{\tabcolsep}{4pt}
        
        \begin{tabular}{l | c c c c c | c} \toprule
            \textbf{Model} & \textbf{Prof.} & \textbf{Pol.} & \textbf{T.O.} & \textbf{Req.} & \textbf{Pref.} & \textbf{Avg} \\
            \midrule[1pt]
            \multicolumn{7}{c}{\textbf{Prompt-Based}} \\ 
            \midrule[1pt]
                   Qwen-2.5-7B-It & 88.3 & 49.2 & 94.3 & 96.0 & 85.7 & 82.7 \\
                  Llama-3.1-8B-It & 90.9 & 41.0 & 97.1 & \underline{99.0} & 81.0 & 81.8 \\
                   Gemma-3-27B-It & \textbf{98.7} & 59.0 & 97.1 & 97.0 & 78.6 & 86.1 \\
                  Qwen-2.5-72B-It & 89.3 & 59.6 & 94.1 & 96.9 & 78.6 & 83.7 \\
                 Llama-3.3-70B-It & \underline{97.4} & \underline{65.6} & \textbf{98.1} & \underline{99.0} & 92.9 & \textbf{90.6} \\
            \midrule[1pt]
            \multicolumn{7}{c}{\textbf{Inference-Time Steering with User Goal}} \\ 
            \midrule[1pt]
                   Qwen-2.5-7B-It & 70.1 & 54.2 & \textbf{98.1} & \underline{99.0} & 90.5 & 82.4 \\
                  Llama-3.1-8B-It & 87.0 & 60.7 & 97.1 & 98.0 & \textbf{100.0} & \underline{88.6} \\
                   Gemma-3-27B-It & 92.2 & \textbf{73.8} & 96.2 & 98.0 & 73.8 & 86.8 \\
                  Qwen-2.5-72B-It & 93.5 & 62.3 & 92.4 & 94.1 & 76.2 & 83.7 \\
                 Llama-3.3-70B-It & \textbf{98.7} & 59.0 & 96.2 & 98.0 & 90.9 & 88.6 \\
            \midrule[1pt]
            \multicolumn{7}{c}{\textbf{Inference-Time Steering with User Goal States}} \\ 
            \midrule[1pt]
                   Qwen-2.5-7B-It & 77.9 & 55.7 & 96.2 & 98.0 & 92.9 & 84.1 \\
                  Llama-3.1-8B-It & 92.2 & 57.4 & 93.3 & 98.0 & 95.2 & 87.2 \\
                   Gemma-3-27B-It & 94.8 & 62.3 & 92.4 & 97.0 & 85.7 & 86.4 \\
                  Qwen-2.5-72B-It & 93.3 & 54.4 & \underline{98.0} & 97.9 & \underline{97.6} & 88.3 \\
                 Llama-3.3-70B-It & 93.5 & 63.9 & \textbf{98.1} & \textbf{100.0} & \underline{97.6} & \textbf{90.6} \\

            \bottomrule
        \end{tabular}
        \caption{$\tau$-Bench Airline}
    \end{subtable}
    \hfill
    \begin{subtable}[t]{0.48\textwidth}
        \centering
        \scriptsize
        \setlength{\tabcolsep}{4pt}
        
        \begin{tabular}{l | c c c c c | c} \toprule
            \textbf{Model} & \textbf{Prof.} & \textbf{Pol.} & \textbf{T.O.} & \textbf{Req.} & \textbf{Pref.} & \textbf{Avg} \\
            \midrule[1pt]
            \multicolumn{7}{c}{\textbf{Prompt-Based}} \\ 
            \midrule[1pt]
                   Qwen-2.5-7B-It & 82.0 & 40.8 & 96.0 & \underline{99.5} & 91.7 & 82.0 \\
                  Llama-3.1-8B-It & 84.8 & 36.8 & 97.1 & 98.9 & 96.3 & 82.8 \\
                   Gemma-3-27B-It & 91.0 & 39.5 & 97.8 & \underline{99.5} & 96.3 & 84.8 \\
                  Qwen-2.5-72B-It & 88.6 & 40.8 & 98.2 & 97.9 & 91.7 & 83.4 \\
                 Llama-3.3-70B-It & 91.7 & 46.1 & \textbf{99.6} & \textbf{100.0} & \textbf{100.0} & 87.5 \\
            \midrule[1pt]
            \multicolumn{7}{c}{\textbf{Inference-Time Steering with User Goal}} \\ 
            \midrule[1pt]
                   Qwen-2.5-7B-It & 73.4 & 38.2 & 97.5 & 98.4 & 95.4 & 80.6 \\
                  Llama-3.1-8B-It & 91.7 & 38.2 & \underline{98.9} & \underline{99.5} & 94.4 & 84.5 \\
                   Gemma-3-27B-It & \textbf{94.5} & \textbf{63.2} & 97.8 & 98.4 & 91.7 & \underline{89.1} \\
                  Qwen-2.5-72B-It & 89.6 & 40.8 & 97.1 & 98.9 & 97.2 & 84.7 \\
                 Llama-3.3-70B-It & 92.4 & 55.3 & 98.2 & \textbf{100.0} & \underline{98.1} & 88.8 \\
            \midrule[1pt]
            \multicolumn{7}{c}{\textbf{Inference-Time Steering with User Goal States}} \\ 
            \midrule[1pt]
                   Qwen-2.5-7B-It & 73.0 & 47.4 & 94.9 & 97.9 & 93.5 & 81.4 \\
                  Llama-3.1-8B-It & 84.8 & 52.6 & 95.3 & 98.9 & 97.2 & 85.8 \\
                   Gemma-3-27B-It & \textbf{94.5} & 56.6 & 97.1 & \underline{99.5} & 89.8 & 87.5 \\
                  Qwen-2.5-72B-It & 85.8 & 48.7 & 97.1 & 98.4 & \underline{98.1} & 85.6 \\
                 Llama-3.3-70B-It & \underline{92.7} & \underline{59.2} & 96.0 & 98.4 & \textbf{100.0} & \textbf{89.3} \\
            \bottomrule
        \end{tabular}
        \caption{$\tau$-Bench Retail}
    \end{subtable}

    \begin{subtable}[t]{0.48\textwidth}
        \centering
        \scriptsize
        \setlength{\tabcolsep}{4pt}
        
        \begin{tabular}{l | c c c c c | c} \toprule
            \textbf{Model} & \textbf{Prof.} & \textbf{Pol.} & \textbf{T.O.} & \textbf{Req.} & \textbf{Pref.} & \textbf{Avg} \\
            \midrule[1pt]
            \multicolumn{7}{c}{\textbf{Prompt-Based}} \\ 
            \midrule[1pt]
                   Qwen-2.5-7B-It & 54.7 & 18.0 & 78.4 & 81.9 & 73.5 & 61.3 \\
                  Llama-3.1-8B-It & 73.6 & 35.0 & 86.1 & 89.0 & 80.9 & 72.9 \\
                   Gemma-3-27B-It & \underline{80.1} & 50.3 & 95.0 & 97.7 & 92.1 & 83.0 \\
                  Qwen-2.5-72B-It & 72.0 & 31.0 & 91.0 & 93.3 & 90.2 & 75.5 \\
                 Llama-3.3-70B-It & 72.0 & 31.0 & \textbf{97.9} & \textbf{98.6} & \textbf{93.2} & \underline{83.3} \\
            \midrule[1pt]
            \multicolumn{7}{c}{\textbf{Inference-Time Steering with User Goal}} \\ 
            \midrule[1pt]
                   Qwen-2.5-7B-It & 49.8 & 28.0 & 77.8 & 83.2 & 69.4 & 61.7 \\
                  Llama-3.1-8B-It & 73.9 & 49.7 & 92.9 & 95.8 & 80.4 & 78.5 \\
                   Gemma-3-27B-It & \textbf{80.8} & 49.0 & 95.9 & \underline{98.3} & 89.4 & 82.7 \\
                  Qwen-2.5-72B-It & 71.7 & 41.7 & 92.7 & 94.9 & \underline{93.1} & 78.8 \\
                 Llama-3.3-70B-It & 73.6 & \underline{60.3} & 94.9 & 95.8 & 91.0 & 83.1 \\
            \midrule[1pt]
            \multicolumn{7}{c}{\textbf{Inference-Time Steering with User Goal States}} \\ 
            \midrule[1pt]
                   Qwen-2.5-7B-It & 43.4 & 26.3 & 88.1 & 89.8 & 71.1 & 63.7 \\
                  Llama-3.1-8B-It & 71.7 & 38.0 & 85.2 & 88.6 & 74.3 & 71.6 \\
                   Gemma-3-27B-It & 72.3 & 45.7 & 87.0 & 89.7 & 74.8 & 73.9 \\
                  Qwen-2.5-72B-It & 71.0 & 46.3 & 94.9 & 96.9 & 92.9 & 80.4 \\
                 Llama-3.3-70B-It & 72.6 & \textbf{64.3} & \underline{97.0} & 97.8 & 91.7 & \textbf{84.7} \\
            \bottomrule
        \end{tabular}
        \caption{MultiWOZ Challenge}
    \end{subtable}
    
    \caption{User simulator goal alignment performance with prompt-based, inference-time steering with user goal and inference-time steering with user goal states. The table shows the average success rates for User Profile (Prof.), User Policy (Pol.), Task Objective (T.O.), Requirements (Req.) and Preferences (Pref.) sub-component categories, and the overall average performance. Highest scores are \textbf{bolded}, and the second-highest scores are \underline{underlined}.}
    \label{tab:tab:inference_time_steering}
\end{table*}

\section{Naturalness and Coherence Evaluation Prompts}
\label{sec:naturalness_and_coherence_evaluation_prompts}

\begin{promptbox}{Naturalness Evaluation Prompt}
\scriptsize
You are an expert evaluator analyzing a user simulator's behavior while they are pursuing a user goal. Your task is to evaluate the naturalness of the user simulator throughout conversation. \\

\# User Goal \\
\{user\_goal\} \\

\# Conversation History \\
\{conversation\_string\} \\

\# Evaluation Task\\
Rate the naturalness of the user responses on a scale from 1 to 5, based on three key dimensions: \\
\begin{enumerate}
    \item **Grammar and coherence**: Grammatically correct sentences that flow logically and coherently.
    \item **Context Relevance** Responses are strongly related to the conversation context and previous turns, as well as the user goal.
    \item **Conversational Style**: Responses are natural, and appropriate for the conversation style.
\end{enumerate}
    
The rating scale is as follows: \\
\begin{itemize}
    \item 5 (Highly Natural): Excels in all three dimensions - perfect grammar/coreference, fully contextually appropriate, and natural conversational
    \item 4 (Mostly Natural): Strong in all dimensions with minor lapses - occasional awkward phrasing or slightly unnatural language
    \item 3 (Moderately Natural): Adequate in all dimensions - functional but noticeably artificial in style or occasional context mismatches
    \item 2 (Somewhat Unnatural): Weak in 1-2 dimensions - grammar errors, loses context, or overly formal/robotic style
    \item 1 (Very Unnatural): Fails multiple dimensions - poor grammar, irrelevant responses, or mechanical/repetitive patterns
\end{itemize}

\begin{verbatim}
# Output Format:
{{
    "naturalness_score": # number from 1 to 5
    "reasoning": # Brief explanation of your evaluation decision
}}
\end{verbatim}

Output a properly formatted JSON response, as specified by the Output Format.
\end{promptbox}

\begin{promptbox}{Coherence Evaluation Prompt}
\scriptsize
You are an expert evaluator analyzing a user simulator's behavior while they are pursuiing a user goal. Your task is to evaluate the coherence of a user simulator in pursuing their goal in a conversation. \\

\# User Goal \\
\{user\_goal\} \\

\# Conversation History \\
\{conversation\_string\} \\

\# Evaluation Task\\
Rate the coherence of the user's dialogue on a scale from 1 to 5, based on three key dimensions: \\
\begin{enumerate}
    \item **Goal Progression**: User consistently works toward their stated goal with appropriate persistence and adaptation
    \item **Topic Continuity**: Responses maintain topical relevance and logical connections between turns
    \item **Response Appropriateness**: User responds appropriately to system prompts, questions, and clarification requests
\end{enumerate}
    
The rating scale is as follows: \\
\begin{itemize}
    \item 5 (Highly Coherent): Clear goal pursuit throughout, all utterances topically connected, appropriate responses to all system turns
    \item 4 (Mostly Coherent): Strong goal focus with minor deviations, well-connected utterances, occasionally misses response opportunities
    \item 3 (Moderately Coherent): Generally pursues goal with some inconsistencies, mostly connected utterances, some inappropriate responses
    \item 2 (Somewhat Incoherent): Weak goal focus with major deviations, frequent topic jumps, often misses or misinterprets prompts
    \item 1 (Very Incoherent): No clear goal pursuit, disconnected responses, fails to engage appropriately with system
\end{itemize}

\begin{verbatim}
# Output Format:
{{
    "coherence_score": # number from 1 to 5
    "reasoning": # Brief explanation of your evaluation decision
}}
\end{verbatim}

Output a properly formatted JSON response, as specified by the Output Format.
\end{promptbox}

\end{document}